# "Draw My-Topics" Toolkit: Draw Desired Topics Fast from Large Volume of Corpus

**Jason Xiaotian Dou   Ni Sun   Xiaojun Zou**
Department of Computer Science
Peking University, Beijing 100871, P.R. China
{douxiaotian, sn96, xiaojunzou}@pku.edu.cn

## Abstract

We develop the "Draw My-Topics" Toolkit, which provides a fast way to incorporate social scientists' concerns and interests into the standard topic model. Instead of using raw corpus with primitive processing as input, an algorithm based on Vector Space Model and Conditional Entropy are used to connect social scientists' subjective want and the unsupervised topic models' output. Space for users' adjustment on specific corpus of their interest is accommodated in our algorithm. We demonstrate the toolkit's use on the Diachronic People's Daily Corpus in Chinese. Several interesting "central words" like "Enlai Zhou" (First PRC premier minister) and "Cultural Revolution" which may be interested of social scientists from different disciplines and the original corpus are used as input of our toolkit, then the most related topics are present efficiently for further research purpose.

## 1   Introduction

Probabilistic topic models, such as Probabilistic Latent Semantic Analysis (PLSA) and Latent Dirichlet Allocation (Blei, Ng and Jordan, 2003), are widely used as common tools to assist social scientists' understanding large, unstructured collections of documents. The value of topic models is being recognized by social scientists as a tool for large document-summary and abstraction for economically and politically interesting facts such as Chinese Censorship (Grimmer and Stewart, 2013; King, Robert and Pan, 2013; Tingley, 2013; Bamman, Connor and Smith, 2012).

Social scientists often start with off-the-shelf implementations of topic modeling which are widely available on the Internet. Then a variety of post-hoc evaluation of the implementation's output, including topic prevalence and topic variation can be conducted. However, due to topic models are mainly unsupervised method. In this case, social scientists often have little to do with the topic generation process. So many unrelated topics may show up, but they are not of the social scientists' interest. There are already considerable terrific works on better connecting social scientists with topic modeling. Kim, Zhai and Diermeier (2013) connected topic modeling with time-series feedback, Roberts, Stewart, Tingley and Airoldi (2013) developed "Structural Topic Model" to incorporate corpus observed metadata into standard topic model. Hall, Wallach, Mimno and McCallum (2009) accommodated outside information by optimize the hyper parameters of LDA. Hall, Jurafsky and Manning (2008) hand selected seed-word by adding number of pseudo-counts to the topic related words that they are especially interested in. We develop the "Draw Related-Topics" toolkit to help social scientists and other topic model users to get desired topics in a more direct way. The central idea is that users define their interesting topics by a "central word", and then we extract this word (topic)'s relatively small context rather than the huge volume of raw corpus as topic model's input. Based on "Spatial Locality Principle", this allows us to draw central word's related topic prevalence and related topics much easier than searching for the whole corpus purposely. To define and find the "related context", we propose a two –step approach. First, to find the "central word's top twenty similar words by Vector Space Model (Salton, Wong and Yang, 1975) and Conditional Entropy (Cover, Thomas M, 1991). These form the similar word set. Second, extract adjacent context of the similar word set to form the whole related context. Furthermore, users can adjust the two approaches by their subjective judgment (in other words, social science sense/knowledge) according to their own corpus' part-of-speech-tagging statistics to get more desirable results. After describing the method, we demonstrate the use of "Draw Related-Topics" toolkit by analyzing several interesting words on the diachronic "People's Daily" corpus in Chinese.

## 2 The Two-Step Approach

The input of our "Draw My-Topics" Toolkit is interesting words defined by users (target at social scientists mainly) and large volume of corpus. The output is "central word" related topics content and topic prevalence. Also, users can adjust the output by their domain knowledge and intuitions by flexible parameters we provide.

### 2.1 Similar Words Calculation

In the first step, we calculate top three hundred similar words of each given "Central Word" by vector space model and conditional entropy. Vector space model is an algebraic model for representing text documents as vectors of identifiers. In our case, each word is treated as a vector in the space. The similarity degree of different words is calculated by the cosine of the angle between different vectors of words. The entry of word vector is point-to-point mutual information. Then to calculate mutual information, decision on length of information window gets crucial and subtle. We do this based on "amount of information" of each window, which is calculated as conditional entropy.

$$\text{Information} = -\log(\Pr(X, Y) / \Pr(Y))$$

In it Y denotes the target word and X denotes words in nearby context. For four part-of-speech tagging types, we set four different information thresholds as the following based on sampling, observation, and statistics:

| Part-of-speech | Noun | Verb | Adjective | Adverb |
|---|---|---|---|---|
| Left-threshold | 21 | 24 | 7 | 12 |
| Right threshold | 14 | 15 | 9 | 20 |

Table 1

This information threshold table for similarity degree calculation can also be determined by toolkit users themselves since "similar" is quite a subjective measure from different disciplines' perspective. For example, "demand" may be "similar" to "supply" from an economist's view while political scientist may think "demand" is related to "power".

Some of the similarity calculation results based on Chinese diachronic corpus of *"People's Daily"* are presented below.

| Target words | 万象 | 人民 | 只求 |
|---|---|---|---|
| Similar words in Decedent Order | 佩差 | 代表大会 | 韩仲琦 |
| | 雅林 | 中国 | 幻觉 |
| | 僧侣 | 共和国 | 胆子 |
| | 冯 | 条例 | 洞房 |
| | 苏发 | 委员会 | 背 |
| | 老挝 | 是 | 憎恶 |
| | 串通一气 | 全国组织 | 独白 |
| | 亲王 | 友谊 | 新娘 |
| | 努 | 主席 | 促使 |
| | 流亡 | 社会主义 | 现实 |

Table 2

### 2.2 Summarize Related Corpus

In the second step, we apply a straightforward method to summarize related corpus from the original one based on similarity words result from first step. We go through the People's Daily Corpus year by year. For every line of one year, draw it down if the line contains any of the similar words of given target word. All these lines constitute our related corpus. In our "People's Daily Corpus", every line is a separate news piece, so this method take the completeness of news well.

Drawback of this kind of summary is obvious: we reduce corpus size as input of topic model at the risk of neglect precious information related to our target word in the abandoned corpus.

Figure 1 shows part of related corpus we draw for central word "Enlai Zhou"(周恩来/n), who is PRC's first premier minister.

福建省/n 黎明/t 农业社/n 去年/t 收入/n 百万/m 元/q 以上/f 广东省/n 梅/g 田/n 农业社/n 社员/n 收入/n 成/v 几/m 倍/q 增加/v 荒/g 弃/g 多年/m 的/u 野/b 茶树/n 又/d 成/v 了/u 有用/a 的/u 财富/n 发展/v 郊区/s 生产/v 增加/v 副食/n 供应/v 　天津/n 郊区/s 去年/t 供应/v 市内/s 的/u 肉类/n 大大/d 增加/v 　上海/n 郊区/s 已/d 能/v 供应/v 全市/n 所/u 需/v ７０％/m 的/u 蔬菜/n 　农业/n 生产/v 受灾/v 减产/v ，/w 用/p 副业/n 生产/v 弥补/v 起来/v 幸福/a 之/u 路/n 社/n 多数/m 社员/n 仍/d 能/v 增加/v 收入/n 两/m 套/q 合/v 一/m 套/q ，/w 争吵/v 变/v 协调/v 　太谷县/n 试行/v 商业/n 体制/n 改革/v 的/u 经过/n 和/c 效果/n 不要/d 夸大/v ，/w 也/d 不/d 要/v 缩小/v 常/d 到/v 社员/n 家里/s 谈谈/v 心/n 作/v 一个/m 道德/n 品质/n 高尚/a 的/u 人/n 四川/n 、/w 阜新/n 、/w 本溪/n 向/p 群众/n 进行/v 共产主义/n 道德/n 教育/v 中共/n 河北/n 省委/n 抽调/v 一/m 批/q 负责/v 干部/n 参加/v 各县/r 实

际/n 领导/n 工作/v 国务院/n 机关/n 事务/n 管理局/n 要求/v 机关/n 团体/n 部队/n 节约/v 烤火/v 用/p 煤/n 历时/v 五十/m 天/q 往返/v 六/m 千/m 里/q 中央/n 慰问团/n 三分/t 团/n 结束/v 慰问/v 捷/g 冰球队/n 昨/t 胜/v 北京队/n 新闻/n 简讯/n 真正/b 的/u 爱情/n 从/p 一/m 句/q 话/n 想到/v 一/m 件/q 事/n 对/p 批评/v 的/u 反应/v 全国/n 唯一/b 的/u 水族/n 聚居区/n 三都水族自治县/n 成立/v 都/d 江/n 上游/f 的/u 欢笑/v 三都水族自治县/n 介绍/v 在/p 越南/n 国民/n 大会/n 第六/m 次/q 会议/n 上/f 范文同/n 总

Figure 1

## 3 Experimental Results and Online Visualization

In this part, we first demonstrate how the "Draw My-Topics" Toolkit can effectively condense corpus size, then we show the topics draw from the condensed corpus, finally, we show our online visualization platform. Online service and downloadable package will all be provided soon.

In the following table, four interesting words (周恩来, Enlai Zhou; 北大, Peking University; 经济, economy; 老天爷, Chinese God) and the whole year 1957's corpus of People's Daily are used as input of our toolkit, we can see that size of related corpus decrease significantly in the four cases.

| target | 周恩来 | 北大 | 经济 | 老天爷 |
|---|---|---|---|---|
| Original size | 69180KB | 69180KB | 69180KB | 69180KB |
| Condensed size | 15937KB | 3622KB | 42636KB | 9024KB |

Table 3

But our goal is topic prevalence and topic content; will the condensed corpus work well? Here are the results in Table 4. Stop words are removed from the corpus to distinguish the four topics.

| Target words | 周恩来 | 北大 | 经济 | 老天爷 |
|---|---|---|---|---|
| Topic Distribution | 人民 | 美国 | 人民 | 人民 |
| | 工作 | 人民 | 国家 | 还 |
| | 社会主义 | 国家 | 工作 | 美国 |
| | 中国 | 问题 | 问题 | 右派 |
| | 国家 | 政府 | 社会主义 | 分子 |
| | 问题 | 苏联 | 中国 | 领导 |
| | 总理 | 英国 | 生产 | 生产 |
| | 政府 | 党 | 美国 | 进 |
| | 主席 | 中国 | 党 | 思想 |
| | 领导 | 社会主义 | 发展 | 群众 |
| | 建设 | 进行 | 苏联 | 干部 |
| | 思想 | 工作 | 方面 | 共产党 |

Table 4

For the convenience of users to use our toolkit and other "products", we are also building an online visualization platform based on the diachronic corpus of "People's Daily". Functions include plot of fifty-year word frequency distribution as showed in Figure 2 and ten years' similarity degree variation of given word as showed in Figure 3. A dynamic visualization of our toolkit's application will be implemented on the platform soon.

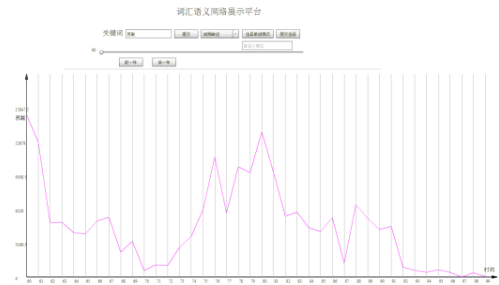

Figure 2

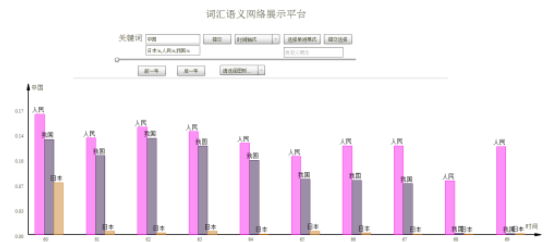

Figure 3

## 4 Conclusion and Future Work

We have developed the "Draw My-Topics" Toolkit for social scientists to incorporate their concerns and interests when using standard topic model. The main method is to use word similarity calculation based on Vector Space Model and Conditional Entropy to "condense" original corpus. The condensed size is also helpful when facing large scale of corpora, where Topic Model training time can be a bottleneck. Space for social scientists to incorporate their own judgments is also provided. An online visualization platform and downloadable package will be released soon.

We are improving this work from two aspects, calculation and evaluation. In the calculation part, we hope to incorporate the diachronic ontology we build on "People's Daily" (Shaoda He, et al. 2013) to improve quality of condensed corpus. Evaluation of topic modeling is quite an open question. Chang, Boyd-Graber, Gerrish, Wang and Blei (2013) use human judgments including user studies to examine the topics. In our case, we hope to use the user feedback of our toolkit to design new feasible methods.

## Acknowledgements

Thanks to Noah Smith, Gary King, Brandon Stewart, Shuo Chen, Xun Pang, David Hall and Sebastian Benthall for enlightening suggestions, comments and guide; and also to my Undergraduate Research advisor, Professor Junfeng Hu. It is his guide, dedication and encouragement that make me fall in love with research and finish my first paper as this. "National Innovation Plan for Undergraduate", Ministry of Education of China supports this work.